\documentclass[11pt]{article}

\usepackage[preprint]{acl}

\usepackage{times}
\usepackage{latexsym}

\usepackage[T1]{fontenc}

\usepackage[utf8]{inputenc}

\usepackage{microtype}

\usepackage{inconsolata}

\usepackage{graphicx}
\usepackage{booktabs}
\usepackage{placeins}
\usepackage{tabularx}
\usepackage{amsmath}
\usepackage{amssymb}
\usepackage{xcolor}
\usepackage{algorithm}
\usepackage{algpseudocode}
\usepackage{multirow}
\usepackage{tikz}
\usetikzlibrary{arrows.meta,positioning,calc}

\newlength{\csw}
\newcommand{\missingfigure}[1]{%
  \fbox{\parbox[c][0.72in][c]{\csw}{\centering\scriptsize Missing\\figure}}%
}
\newcommand{\maybeincludegraphics}[2][]{%
  \IfFileExists{#2}{\includegraphics[#1]{#2}}{\missingfigure{#2}}%
}
\algrenewcommand\algorithmicrequire{\textbf{Require:}}
\algrenewcommand\algorithmicensure{\textbf{Ensure:}}
\tikzset{wmflowbox/.style={rectangle, draw=black!55, rounded corners=1pt, fill=blue!4, thick, inner sep=4pt, align=center, font=\scriptsize}}

\title{Mind-Studio: Executable World Models with Lookahead Evaluation for Partially Observable Games}
\makeatletter
\newcommand{\emailgrp}[2]{%
  \def\emailgrp@sep{}%
  \@for\emailuser:=#1\do{%
    \emailgrp@sep\def\emailgrp@sep{\quad}%
    \texttt{\emailuser @#2}%
  }%
}
\makeatother

\author{%
  \textbf{Yifei Dong}$^{1}$,
  \textbf{Mingen Zheng}$^{1}$,
  \textbf{Linquan Wu}$^{2}$,
  \textbf{Jeff Z. Pan}$^{3}$,
  \textbf{Jiaxin Bai}$^{4}$
  \\[0.25em]
  \normalfont
  $^{1}$Hong Kong University of Science and Technology \quad \\
  $^{2}$City University of Hong Kong
  $^{3}$University of Edinburgh 
  $^{4}$Hong Kong Baptist University \\
  \normalfont
  \emailgrp{ydongbl, mzhengap}{connect.ust.hk} \quad \\
  \texttt{Linquan.Wu@my.cityu.edu.hk} \quad
  \texttt{j.z.pan@ed.ac.uk} \quad
  \texttt{baijiaxin@hkbu.edu.hk} 
  \\ \url{https://github.com/HKBU-KnowComp/MindStudio}
}

\begin{document}
\maketitle

\begin{abstract}
World-model synthesis aims to turn interaction experience into an internal model of environment dynamics. Existing symbolic approaches often fit observed transitions or mixtures of local rules, but they do not produce a complete executable program that can run independently of the real environment. We present \textbf{Mind-Studio}, a framework that synthesizes executable \texttt{pygame}-style world models from state-action-next-state trajectories using large language models. Mind-Studio combines entropy-selected traces with a lightweight game skill file containing object, action, and static scene information extracted from screenshots. We evaluate synthesis quality with a $K$-step lookahead fidelity protocol that compares generated WM rollouts against Real-ALE rollouts from the same state. On \emph{Montezuma's Revenge}, Mind-Studio improves chosen-action NSP from $0.3\%$ (PoE-World) to $48.7\%$ while verifying $5/8$ subgoals; across \emph{Alien}, \emph{Assault}, and \emph{Skiing}, it achieves stronger branch-level fidelity than prior learned lookahead sources.
\end{abstract}

\section{Introduction}

\begin{figure*}[!t]
  \centering
  \setlength{\tabcolsep}{3pt}
  \setlength{\csw}{0.23\textwidth}
  \setlength{\fboxsep}{0pt}
  \setlength{\fboxrule}{0.4pt}
  \begin{tabular}{@{}cccc@{}}
    \fbox{\includegraphics[width=\csw]{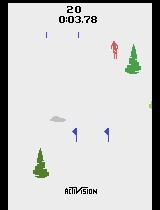}} &
    \fbox{\includegraphics[width=\csw]{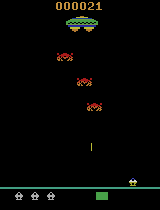}} &
    \fbox{\includegraphics[width=\csw]{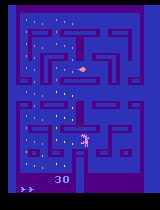}} &
    \fbox{\includegraphics[width=\csw]{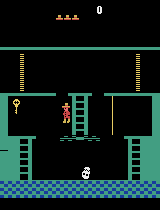}} \\
    \multicolumn{4}{c}{Original game-state renderings} \\[2pt]
    \fbox{\includegraphics[width=\csw]{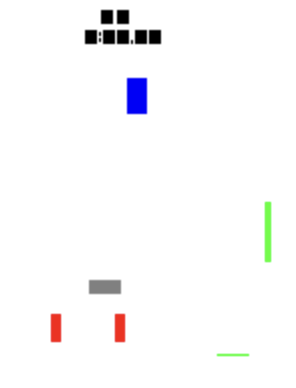}} &
    \fbox{\includegraphics[width=\csw]{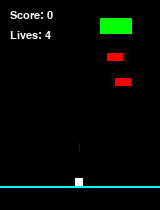}} &
    \fbox{\includegraphics[width=\csw]{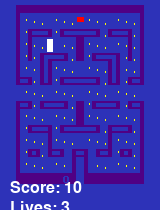}} &
    \fbox{\includegraphics[width=\csw]{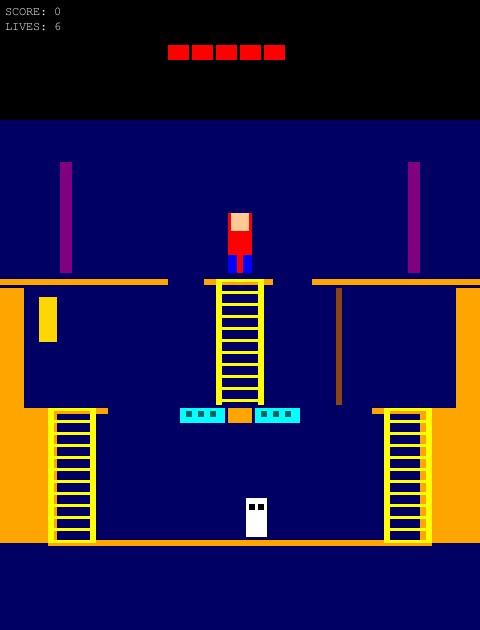}} \\
    \multicolumn{4}{c}{Mind-Studio game renderings} \\[2pt]
    Skiing & Assault & Alien & Montezuma
  \end{tabular}
  \caption{\textbf{Representative game renderings.} Top: original Atari game-state renderings. Bottom: Mind-Studio renderings produced by the generated code-based world models, showing the abstract object-centric representation.}
  \label{fig:v5_case_studies}
\end{figure*}

A world model is an internal representation that describes how an environment changes under actions, allowing an agent to reason beyond the current observation \cite{ha2018worldmodels,sutton1998reinforcement}. For decision making, such a model should not merely predict the next state once; it should encode enough dynamics to support hypothetical rollouts under different actions.

Symbolic world models are a natural fit for this goal because they represent environments through objects, attributes, relations, and rules \cite{garcez2019neural}. Code-based symbolic world models go one step further: object states become program variables, while dynamics are expressed through control flow, functions, and numerical updates \cite{pddl}. 

Recent LLM-based systems show that programmatic world models can be induced from state-action-next-state tuples \cite{worldcoder,poe}. However, many existing methods remain transition-centric: they reproduce observed input-output behavior or combine local rule experts, but do not necessarily synthesize a complete mechanism that explains how the transition is produced. In partially observable Atari games, this distinction matters. The same visible state can hide different internal conditions, and multiple mechanisms---movement, collision priority, enemy behavior, respawn, and object collection---often interact in the same local region.

This motivates the central question of this work: how can we synthesize an executable world model that captures these mechanisms as a runnable program, rather than only fitting surface-level transitions? A useful world model should preserve object identities, static scene constraints, action semantics, and update order, so that its behavior remains coherent when executed from different states.

We propose \textbf{Mind-Studio}, a code-based symbolic world-model synthesis framework. Mind-Studio builds symbolic trajectories from pixel- or RAM-derived OC-Atari objects, selects informative transitions using entropy-based scoring, compresses the selected traces into static interfaces and action-conditioned deltas, and prompts an LLM to generate a \texttt{pygame}-style transition program guided by a compact game skill file. The resulting standalone program is the main artifact: it can update object states, render predicted scenes, and expose the rules used to produce each transition.

We evaluate the synthesized WMs on \emph{Montezuma's Revenge}, \emph{Alien}, \emph{Assault}, and \emph{Skiing} (Figure~\ref{fig:v5_case_studies}). These games stress different synthesis challenges, including hidden boundary conditions, moving enemies, stochastic obstacles, and jump-sensitive interactions. We measure fidelity with a $K$-step lookahead protocol that compares WM and Real-ALE endpoints from the same snapshot. On Montezuma, Mind-Studio improves chosen-action NSP from $0.3\%$ (PoE-World) to $48.7\%$ while reaching $5/8$ subgoals. Across the three stochastic games, Table~\ref{tab:main_results} shows that Mind-Studio generally produces stronger executable WMs, with especially clear branch-fidelity gains on Assault and Alien and competitive results on Skiing, demonstrating the effectiveness of the synthesized executable WMs.

\paragraph{Contributions.}
Mind-Studio makes four contributions: \textbf{(i)} a single-pass LLM synthesis pipeline guided by a game skill file that encodes static layout, action semantics, and domain conventions---the skill file can be populated from pixel-level scene extraction, enabling use even when RAM-based state is unavailable; \textbf{(ii)} an entropy-based selector for transitions that expose hidden-variable and context-priority rules; \textbf{(iii)} an object-centric program representation with explicit priority ordering that encodes hidden rules as executable control flow; and \textbf{(iv)} a $K$-step lookahead evaluation protocol with a shared LLM-as-Planner harness for controlled comparisons across lookahead sources and planner backbones.All code, synthesized programs, and planning logs are publicly available.

\section{Related Work}
\begin{figure*}[!t]
  \centering
  \includegraphics[width=\textwidth]{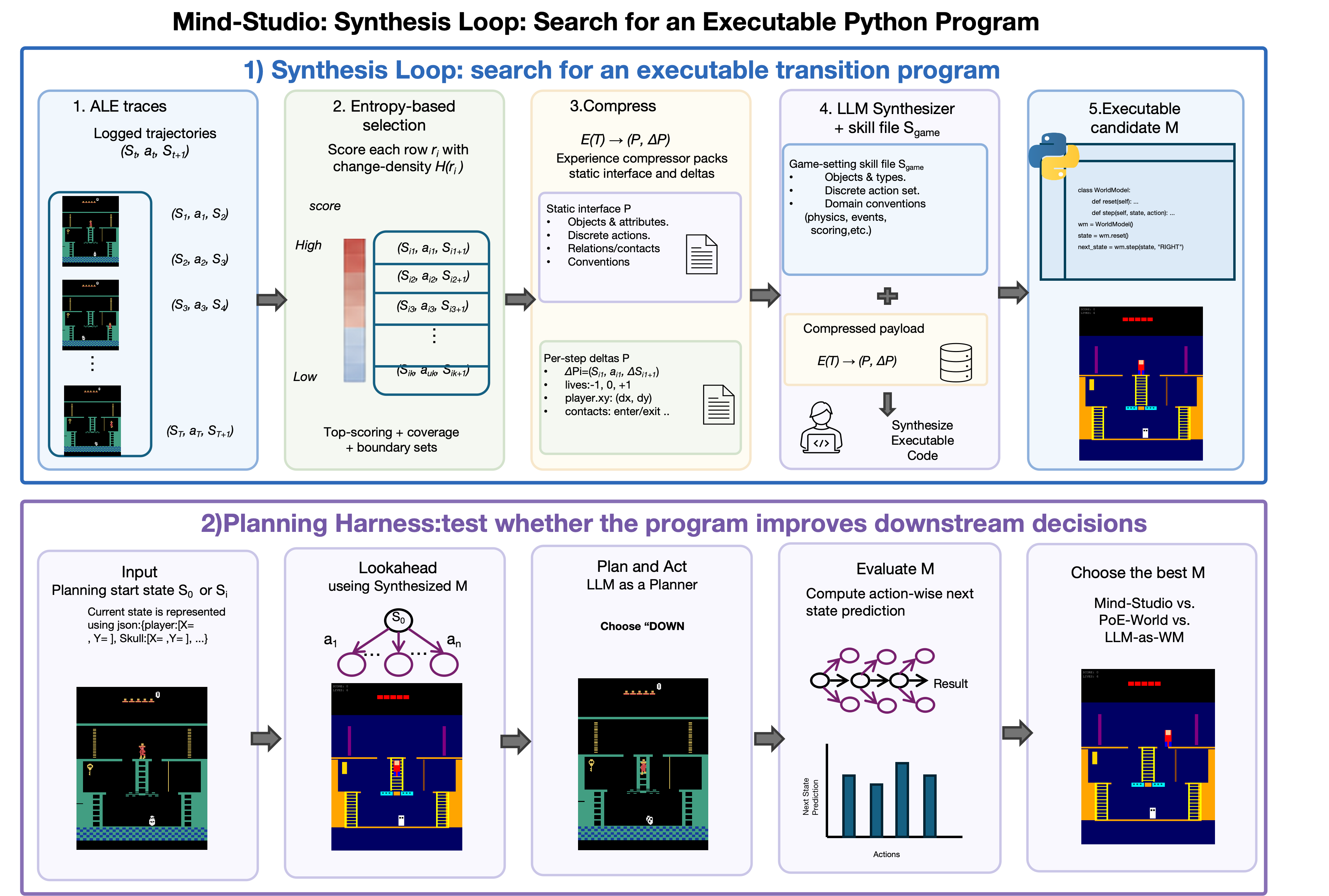}
  \caption{\textbf{Mind-Studio WM synthesis pipeline.} Entropy selection picks rule-bearing rows $\widetilde{\mathcal{T}}$; compressor $E(\cdot)$ packs static interface $\mathcal{P}$ and per-step deltas $\Delta\mathcal{P}$; the LLM emits executable $M$ guided by the game-setting skill file.}
  \label{fig:wm_synthesis_flow}
\end{figure*}

\paragraph{World models for LLM planning.}
World models let agents simulate action consequences before committing \cite{ha2018worldmodels,sutton1998reinforcement}.
Recent LLM-based approaches use the model itself for tree-search planning \cite{rap}, text-state simulation \cite{textsimulators}, implicit next-state prediction \cite{word2world}, retrieval-augmented imagination \cite{rawm}, web-policy co-evolution \cite{webevolver,coex}, neurosymbolic MPC-style rollouts \cite{walle}, or planner/world-model separation \cite{swap}.
Mind-Studio follows this goal of inference-time lookahead, but differs by asking for one executable, object-centric transition program whose $K$-step branch endpoints can be rendered and checked pixel-wise against Real-ALE \cite{bellemare2013arcade} rollouts, rather than free-form language observations or many weighted rules.

\paragraph{Programmatic and compositional world models.}
\label{sec:rw_mdp}\label{sec:rw_pomdp}
Program synthesis offers inspectable, reusable dynamics \cite{ellis2021dreamcoder,garcez2019neural}; WorldCoder and GIF-MCTS synthesize Python transition code through online exploration or MCTS-guided offline search \cite{wang2024worldcoder,dainese2024generating}, while REPL-Plan and FPWC use executable feedback or refinable graph world models for planning \cite{replplan,fpwc}, and PDDLGym provides symbolic planning tasks in a gym interface \cite{pddl}.
Text2World benchmarks LLM-based symbolic world model generation from text \cite{text2world2025}, and \citet{khan2026one} infer symbolic world models for stochastic environments from unguided exploration---both works share our goal of turning interaction experience into reusable transition programs.
Mind-Studio shares this propose--execute--verify spirit but targets OC-Atari \cite{ocatari} POMDPs \cite{kaelbling1998planning} where overlapping contacts, stochastic respawns, and hidden variables require explicit mechanism priority in a single control-flow program---in contrast to PoE-World's product of programmatic experts with learned weights \cite{piriyakulkij2025poeworld} or AgentOWL's hierarchical learned options \cite{agentowl}.

\section{Problem Formulation}
\label{sec:method}
\paragraph{Three roles.}
Mind-Studio involves three roles: (1)~the \textbf{world model} $M$, an executable Python program that predicts how the game state changes after an action; (2)~the \textbf{player}, the controllable character in the Atari game environment, whose screen coordinate is the primary signal for evaluation; and (3)~the \textbf{LLM planner} $\pi_\text{LLM}$, a language model that selects actions by consulting per-action previews produced by $M$.

Let $\mathcal{D}=\{(o_t,a_t,o_{t+1})\}$ be logged transitions from a partially
observed game, with each raw observation $o_t$ converted into an
\emph{object-centric symbolic state} $s_t$---a structured list of game objects
(player, enemies, items) each represented by typed attribute slots such as
position, size, and status, extracted via the OC-Atari RAM
parser~\cite{ocatari}. The goal is to synthesize an executable world model $M$
that rolls this symbolic state forward under an action:
\begin{equation}
  \hat{s}_{t+1}=M(s_t,a_t).
  \label{eq:wm-forward}
\end{equation}
We evaluate $M$ both as a replay predictor and as the lookahead source for a
real-game planner. Montezuma progress is an ordered subgoal sequence
$\{G_1,\ldots,G_J\}$; the other games use achievement counts such as gates,
kills, or eggs. Table~\ref{tab:symbols} summarises notation used throughout.

\paragraph{World model $M$.}
$M$ is a \texttt{pygame}-style Python transition program. Its static interface $\mathcal{P}$ holds persistent object slots; its transition code updates positions, contacts, deaths, and spawns. The predicted state can be rendered so both the planner and evaluator inspect the same branch previews.

\begin{table}[!t]
  \centering
  \scriptsize
  \setlength{\tabcolsep}{4pt}
  \renewcommand{\arraystretch}{1.05}
  \begin{tabular}{@{}ll@{}}
    \toprule
    \textbf{Symbol} & \textbf{Meaning} \\
    \midrule
    $\mathcal{D}$ & Logged replay transitions $\{(o_t,a_t,o_{t+1})\}$ \\
    $s_t$ & Object-centric symbolic state at step $t$ \\
    $M$ & Synthesized executable world model (\texttt{pygame}-style) \\
    $M^\star$ & Best (highest-NSP) program after optional refinement \\
    $\mathcal{P}$ & Static game interface held in $M$ \\
    $\Delta\mathcal{P}$ & Per-step object deltas (compact experience format) \\
    $\widetilde{\mathcal{T}}$ & Entropy-selected training rows \\
    $\hat{s}_{t+1}$ & WM-predicted next state \\
    $K$ & Lookahead horizon (frames per candidate action) \\
    $\pi_\text{LLM}$ & LLM planner selecting actions via $K$-step previews \\
    $\text{NSP}$ & Next-state prediction hit rate (player \texttt{xy} exact match) \\
    $P$ & Task-progress score (kills / subgoals / gates reached) \\
    $\Delta P$ & Progress increment of a planning session \\
    \bottomrule
  \end{tabular}
  \caption{\textbf{Notation summary.}}
  \label{tab:symbols}
\end{table}

\paragraph{Evaluation axes.}
We report two complementary outcomes. \emph{Next-state prediction (NSP)} measures
whether model previews match Real-ALE rollouts; \emph{success rate (SR)} measures
whether those previews help a planner make real task progress. We distinguish
\emph{offline NSP} (fixed held-out states, planner not in the loop) from
\emph{online NSP} (states actually visited during closed-loop planning); both use
the same $K$-frame hold protocol defined below.

At each planning round $r$, the harness starts from the same ALE snapshot and
evaluates every candidate action $a\in\mathcal{A}_r$. The action is held for $K$
frames under both $M$ and Real ALE. A single ALE frame rarely produces a
measurable player displacement---for instance, a jump in Montezuma's Revenge
requires \textsc{right} followed by 7 \textsc{noop} frames before the player
reaches peak height---so $K$ is set to the smallest value in $\{1,4,8,16\}$
that maximises held-out NSP for each game, yielding $K{=}16$ for Skiing and
$K{=}8$ for the remaining three. This is an \emph{action-granularity hold}, not
a multi-step planner: for each candidate $a$ we roll out exactly $K$ frames of
the same action to obtain a comparable endpoint, rather than composing distinct
actions over multiple decision steps.
A branch is correct when the two rollouts end at the same extracted player position:
\begin{equation}
  h_r(a)=\mathbf{1}\!\left[\phi(M^K(s_r,a))=\phi(R^K(s_r,a))\right],
  \label{eq:nsp-branch}
\end{equation}
where $R^K$ denotes the Real-ALE $K$-frame hold rollout and $\phi(\cdot)$
extracts the player's screen coordinate---the pixel-level $(x,y)$ position of
the controllable character's bounding-box centre. We use the player coordinate
as the evaluation signal because objects on fixed patrol routes (enemies, items)
are easier to predict and are already supplied to the planner as part of the
predicted state $\hat{s}$; the player position is the key decision-relevant
quantity that varies with the chosen action. We summarize planning-time fidelity
in two ways:
\begin{equation}
\begin{aligned}
  \mathrm{NSP}_{\mathrm{branch}}
    &= \frac{\sum_{r,a} h_r(a)}{\sum_r|\mathcal{A}_r|}, \\
  \mathrm{NSP}_{\mathrm{chosen}}
    &= \frac{\sum_r h_r(a_r^\star)}{|\mathcal{R}|}.
\end{aligned}
\label{eq:nsp-summary}
\end{equation}
Here $\mathcal{R}$ is the set of comparable planning rounds. Branch NSP averages
over all previews shown to the planner; chosen-action NSP averages only over the
committed action $a_r^\star$. Per-action NSP is kept separate for diagnosis
(Table~\ref{tab:action_precision}); it is not supplied to the planner as a
hand-designed scalar action score.

Before closed-loop planning, synthesis selects $M^\star$ by one-step replay NSP on held-out tuples: the fraction of $(s,a,s')\in\mathcal{D}_a$ where $\phi(M(s,a))=\phi(s')$, averaged over all actions.

\paragraph{Success rate (SR).}
For Montezuma, SR is the fraction of ordered subgoals verified in the \emph{real}
environment within the step budget:
\begin{equation}
  \mathrm{SR} \;=\; m^{\star}/J,
  \label{eq:sr}
\end{equation}
where $m^{\star}$ is the largest verified subgoal index. For non-Montezuma games
we report the game-specific achievement count directly.

\paragraph{Planning interface.}
The LLM planner $\pi_{\text{LLM}}$ receives three inputs: the current symbolic
state $s_t$, the verified subgoal prefix $m$, and a compact table of per-action
world-model previews,
$B_t=\{M^{K}(s_t,a)\}_{a\in\mathcal{A}}$. It then outputs the next action, or a
short action chunk:
\begin{equation}
  a_t \sim \pi_{\text{LLM}}\!\left(\cdot \mid s_t,B_t,m\right).
  \label{eq:llm-policy}
\end{equation}
The table contains the serialized predicted object states and their rendered
endpoints. The subgoal predicates $\Phi_k$ are not passed as binary action scores
to the LLM; they are used to update the verified prefix only after the committed
action is executed in the real environment.
Terminal resets or $N_{\text{stuck}}$ non-progress steps trigger rollback to the last verified checkpoint (Algorithm~\ref{alg:llm-planner}).

\section{The Mind-Studio Framework}
\label{sec:mindstudio}

Mind-Studio has two stages (Figure~\ref{fig:wm_synthesis_flow}). In the
\textbf{synthesis stage}, it (i)~selects rule-bearing transitions via entropy
scoring, (ii)~compresses them into a base-state and state-increment format, and
(iii)~calls an LLM synthesizer to emit an executable world model $M$, which is
then (iv)~verified on held-out replay tuples to select $M^\star$. In the
\textbf{planning stage}, $M^\star$ is plugged into a shared LLM-as-Planner
harness, so performance differences across WM sources reflect lookahead fidelity
rather than a different planner implementation.

\subsection{World Model Synthesis}
\label{sec:wm_synthesis}
Let $\mathcal{T}=\{(s_t,a_t,s_{t+1})\}$ be the symbolic transitions extracted
from $\mathcal{D}$. We select a rule-bearing subset
$\widetilde{\mathcal{T}}\subset\mathcal{T}$, compress it, and give the result to
an LLM synthesizer together with a game-setting skill file
$\mathcal{S}_{\text{game}}$:
\begin{equation}
  M_{\text{cand}} \;=\; g_\phi\!\big(\,\mathcal{S}_{\text{game}},\; E(\widetilde{\mathcal{T}})\,\big),
  \label{eq:llm-gen}
\end{equation}
where $g_\phi$ is the LLM synthesizer that produces the world model $M$ as an executable Python program; it is frozen at API inference (no fine-tuning). The skill file $\mathcal{S}_{\text{game}}$ encodes object types, legal actions, domain conventions, and static scene layout (e.g.\ maze wall coordinates for Montezuma). Static layout entries that are absent from RAM-derived state can be obtained by pixel-level scene extraction, making the approach applicable to games where only visual observations are available. Beyond geometric layout, the skill file can include high-level game background and goal descriptions that help the LLM orient synthesis without relying solely on interaction traces. The compressor $E(\cdot)$ converts $\widetilde{\mathcal{T}}$ into a prompt-friendly format that maximises information per token: it records the static interface $\mathcal{P}$ (base state: persistent object slots shared across all steps) and per-step deltas $\Delta\mathcal{P}_i=(s_{t_i},a_{t_i},\delta s_{t_i})$ (state increments: only the attributes that changed at step $i$), such as \texttt{player.xy} displacement, \texttt{lives} decrements, and contact onsets or offsets. Together, $\mathcal{P}$ and $\{\Delta\mathcal{P}_i\}$ give the LLM synthesizer $g_\phi$ the persistent game interface and the rule-bearing transitions it needs to emit $M$. The synthesis target is an executable Python program with a \texttt{pygame}-style simulator interface (including state transition and rendering hooks). Each candidate is executed on a held-out replay split; the program with the highest $\mathrm{NSP}_{\text{replay}}$ (Eq.~\eqref{eq:nsp-summary}) becomes $M^{\star}$ for planning.

\textbf{Entropy-based selection of $\widetilde{\mathcal{T}}$.}
\label{sec:entropy_selection}
Most transitions repeat common dynamics, while rare rows reveal contacts,
spawns, and context-dependent controls. The goal of the heuristic $H(r_i)$ is to
maximise the information density of the synthesizer's context: by upweighting
rows that exhibit object-attribute changes, relational transitions, and discrete
game events, the selector covers diverse game mechanics rather than spending the
LLM context budget on repeated movement. We therefore score each row
$r_i=(s_t,a_t,s_{t+1})$ with $H(r_i)$ and prefer rows where informative state
changes occur. The same core score is used for all four
games; Montezuma additionally forces coverage for rare interaction types such as
\texttt{Rope} (a climbable vertical rope that changes player movement physics: the
player swings rather than walks) and \texttt{Conveyer\_Belt} (a platform surface
that applies a constant horizontal drift to the player):
\begin{equation}
\begin{aligned}
  H(r_i) \;=\;
  &\sum_{o \in \Delta^{\text{obj}}_i}
    \omega_{\text{change}}(o) \\
  &+ \omega_{\text{rel}}
    \big|R_i^{\text{after}} \triangle R_i^{\text{before}}\big| \\
  &+ \omega_{\text{evt}} \cdot
    \mathbf{1}[\text{event}_i \neq \emptyset],
\end{aligned}
  \label{eq:entropy-score}
\end{equation}
Here $\Delta^{\text{obj}}_i$ lists objects whose attributes changed.
$\omega_{\text{change}}$ upweights appearances and disappearances relative to
ordinary movement; for displacement $(dx,dy)$ we use
$\omega_{\text{move}}(1+\alpha|dx|+\beta|dy|)$. The relation term compares
before/after sets of \texttt{(relation, object\_type)} pairs, and the event term
captures explicit game events. We assemble $\widetilde{\mathcal{T}}$ from the
top-scoring rows plus small coverage and boundary sets; Appendix~\ref{sec:app_impl}
lists the entropy-score constants, and Sec.~\ref{sec:ablation} tests this choice
against a chronological prefix.

We evaluate the synthesized $M^{\star}$ against PoE-World and LLM-as-WM on both NSP (Sec.~\ref{sec:poe_compare}) and SR (Sec.~\ref{sec:planning_results}).

\subsection{LLM-as-Planner: A Stable Evaluation Harness}
\label{sec:planning}\label{sec:planner}

The second stage asks whether the synthesized model improves action selection.
We use a shared LLM-as-Planner harness with subgoal rollback
(Eq.~\eqref{eq:llm-policy}, Algorithm~\ref{alg:llm-planner}) to obtain a
tractable common interface: in our setup, PoE-World's original BFS planner does
not finish a Montezuma episode within an one-hour planning budget. Every
lookahead source produces the same kind of $K$-step branch table, and the
planner commits an action or short action chunk from that table. Thus the
PoE-World numbers below describe its lookahead previews under this harness,
rather than the full native PoE-World planner stack.

The harness tracks the verified subgoal prefix $m$ in the real environment. It
advances from $m$ to $m{+}1$ only when predicate $\Phi_{m+1}$ fires on a real
committed state, never on a prediction. Terminal resets or
$N_{\text{stuck}}$ non-progress steps restore the last verified checkpoint. This
separates the \emph{WM axis}, where the planner is fixed and only the lookahead
source changes, from the \emph{backbone axis}, where $M^\star$ is fixed and the
planner LLM changes.

\begin{table*}[t]
  \centering
  \small
  \setlength{\tabcolsep}{2.5pt}
  \resizebox{\textwidth}{!}{%
  \begin{tabular}{@{}llcccccccccc@{}}
    \toprule
    Game & Planner
      & \multicolumn{4}{c}{Task outcome}
      & \multicolumn{3}{c}{Chosen-action NSP (exact)}
      & \multicolumn{3}{c}{Branch NSP (exact)} \\
    \cmidrule(lr){3-6}\cmidrule(lr){7-9}\cmidrule(lr){10-12}
     & & \textsc{Real} & \shortstack{LLM-as\\WM} & \shortstack{PoE-\\World}
     & \shortstack{\textbf{Mind-Studio}\\\textbf{(ours)}}
     & \shortstack{LLM-as\\WM} & \shortstack{PoE-\\World}
     & \shortstack{\textbf{Mind-Studio}\\\textbf{(ours)}}
     & \shortstack{LLM-as\\WM} & \shortstack{PoE-\\World}
     & \shortstack{\textbf{Mind-Studio}\\\textbf{(ours)}} \\
    \midrule
    \multirow{5}{*}{Skiing}
      & Qwen3-Coder-480B    & 0  & 0  & \textbf{1}  & \textbf{1}  & 0.00 & \textbf{0.17} & 0.09 & 0.13 & \textbf{0.17} & 0.15 \\
      & DeepSeek-V4-Flash   & 0  & 0  & 0           & 0  & 0.15 & \textbf{1.00} & \textbf{1.00} & 0.05 & \textbf{0.72} & \textbf{0.72} \\
      & Llama-4-Scout-17B   & 0  & 0  & 0           & 0  & 0.00 & 0.08 & \textbf{0.15} & 0.05 & 0.08 & \textbf{0.15} \\
      & Claude-4-Sonnet     & 0  & 0  & 0           & 0  & 0.00 & 0.08 & \textbf{0.15} & 0.00 & 0.08 & \textbf{0.15} \\
      & Gemini-2.5-Flash    & 0  & 0  & 0           & 0  & 0.00 & \textbf{0.23} & \textbf{0.23} & 0.00 & \textbf{0.15} & \textbf{0.15} \\
    \midrule
    \multirow{5}{*}{Assault}
      & Qwen3-Coder-480B    & 21 & 63 & 42 & \textbf{84} & 0.03 & 0.68 & \textbf{0.95} & 0.00 & 0.44 & \textbf{0.87} \\
      & DeepSeek-V4-Flash   & 42 & 21 & \textbf{63} & 42          & 0.03 & 0.71 & \textbf{0.88} & 0.01 & 0.33 & \textbf{0.79} \\
      & Llama-4-Scout-17B   & 0  & 0  & 21 & \textbf{84} & 0.00 & 0.26 & \textbf{0.97} & 0.00 & 0.34 & \textbf{0.98} \\
      & Claude-4-Sonnet     & 63 & \textbf{63} & 0  & \textbf{63} & 0.08 & 0.47 & \textbf{0.89} & 0.11 & 0.33 & \textbf{0.91} \\
      & Gemini-2.5-Flash    & 0  & 0  & \textbf{63} & 42          & 0.27 & 0.39 & \textbf{0.95} & 0.23 & 0.36 & \textbf{0.91} \\
    \midrule
    \multirow{5}{*}{Alien\textsuperscript{$\ddagger$}}
      & Qwen3-Coder-480B    & 3        & \textbf{3}        & 0        & \textbf{3}          & 0.00 & 0.48 & \textbf{0.50} & 0.08 & 0.28 & \textbf{0.38} \\
      & DeepSeek-V4-Flash   & 17       & \textbf{13}       & 0        & 3                   & 0.00 & 0.00 & \textbf{0.79} & 0.01 & 0.32 & \textbf{0.49} \\
      & Llama-4-Scout-17B   & 11       & 3                 & 3        & \textbf{20}         & 0.00 & \textbf{0.58} & 0.24 & 0.00 & \textbf{0.39} & 0.20 \\
      & Claude-4-Sonnet     & 38       & 5                 & \textbf{21} & 4                & 0.00 & 0.16 & \textbf{0.55} & 0.00 & 0.16 & \textbf{0.38} \\
      & Gemini-2.5-Flash    & 12       & \textbf{23}       & 13       & 18                  & 0.00 & 0.21 & \textbf{0.26} & 0.00 & 0.22 & \textbf{0.24} \\
    \midrule
    \multirow{5}{*}{Montezuma}
      & Qwen3-Coder-480B    & 5/8 & \textbf{5}/8 & \textbf{5}/8 & \textbf{5}/8 & 0.02 & 0.00 & \textbf{0.49} & \textbf{0.31} & 0.12 & 0.25 \\
      & DeepSeek-V4-Flash   & 5/8 & \textbf{2}/8 & \textbf{2}/8 & \textbf{2}/8 & 0.00 & \textbf{0.11} & 0.00 & 0.00 & \textbf{0.06} & 0.00 \\
      & Llama-4-Scout-17B   & 4/8 & \textbf{5}/8 & \textbf{5}/8 & \textbf{5}/8 & 0.00 & 0.07 & \textbf{0.49} & \textbf{0.25} & 0.09 & 0.25 \\
      & Claude-4-Sonnet     & 8/8 & \textbf{5}/8 & \textbf{5}/8 & \textbf{5}/8 & 0.29 & 0.13 & \textbf{0.66} & \textbf{0.30} & 0.07 & 0.20 \\
      & Gemini-2.5-Flash    & 8/8 & \textbf{8}/8 & \textbf{8}/8 & \textbf{8}/8 & 0.00 & 0.12 & \textbf{0.42} & 0.00 & 0.12 & \textbf{0.23} \\
    \bottomrule
  \end{tabular}
  }
  \caption{\textbf{Online fidelity and task outcome.} Each row fixes the planner LLM and compares lookahead sources. Task columns report the game score; NSP columns report exact-pixel lookahead accuracy. \textsuperscript{$\ddagger$}Alien branch NSP omits \textsc{noop}/\textsc{fire}. Columns place LLM-as-WM next to the \textsc{Real} reference and Mind-Studio (ours) at the right. Bold is computed only over learned lookahead sources (LLM-as-WM, PoE-World, and Mind-Studio), excluding the \textsc{Real} reference; all learned-source ties for the largest value within a row and metric are bolded.}
  \label{tab:main_results}
\end{table*}
\section{Experiments}
\label{sec:experiments}

The experiments ask three questions: whether Mind-Studio gives more faithful planning-time lookahead than LLM-as-WM and PoE-World, which actions account for the gains, and whether the conclusions persist across planner backbones. Each row in Table~\ref{tab:main_results} is an independent rollout, so we interpret NSP--success links as correlational rather than paired causal effects.

\paragraph{Tasks and baselines.}
We use four OC-Atari games~\cite{ocatari} selected to span diverse WM synthesis challenges.
\emph{Montezuma's Revenge} is a maze game with many boundary conditions: platform edges, ladder entry points, rope transfer geometry, and skull-collision distances are all absent from RAM and must be inferred from interaction traces or supplied via pixel-level scene extraction in the skill file; the skull patrol is stochastic and unpredictable.
\emph{Assault} features a player cannon at the bottom and enemy ships that weave left and right to evade fire, requiring the WM to model dynamic evasion rather than fixed trajectories.
\emph{Alien} and \emph{Skiing} stress stochastic enemy/gate spawning and velocity-based player physics, respectively.
Together the four games cover hidden-variable boundary conditions, dynamic opponent behavior, stochastic respawning, and velocity accumulation---distinct synthesis challenges within the OC-Atari object-centric framework.
We compare with \textbf{LLM-as-WM}, which directly predicts next states from the planner LLM, and \textbf{PoE-World}~\cite{poe}, an ensemble-style programmatic WM for partial observability.
PoE-World's native BFS planner did not finish within an eight-hour budget in our setup, so every lookahead source is evaluated under the same LLM-as-Planner harness (Sec.~\ref{sec:planner}); only the preview source changes.

\paragraph{Models and selection.}
For each game, Mind-Studio synthesizes executable candidates from entropy-selected transitions and the game-setting skill file, optionally augmented with induced static-map/rule summaries (Appendix~\ref{sec:app_impl}), then selects the candidate with the best held-out replay NSP (Sec.~\ref{sec:wm_synthesis}); Sec.~\ref{sec:ablation} ablates this input choice.
An optional refinement loop can fold replay failures back into the skill file and regenerate; however, gains were inconsistent across synthesizer models and the additional wall-time was not justified, so all planning results use single-round synthesis.
The Qwen3-Coder rows provide the main head-to-head comparison, using \textsc{Qwen3-Coder-480B-A35B-Instruct-Turbo} as synthesizer unless stated otherwise.
The remaining rows reuse the same selected Mind-Studio programs with four non-Qwen planners, hide generated code from the planner, and score all previews against Real-ALE rollouts, separating program quality from planner formatting effects.
All decoding uses temperature $0$; identifiers, budgets, retry caps, and implementation details are in Appendix~\ref{sec:app_impl}.

\subsection{Model Fidelity}
\label{sec:poe_compare}

\paragraph{Offline fidelity.}
Offline evaluation removes the planner and scores WM rollouts from fixed states on two data splits. The \emph{replay} split contains the trajectories used to synthesize the WM, replayed under the same initial states (Table~\ref{tab:offline_replay_nsp}); the \emph{random} split is a random-policy holdout session not used for synthesis (Table~\ref{tab:offline_random_nsp}). Both tables report chosen-action NSP for the logged action and branch NSP pooled over candidate actions. Mind-Studio leads on random holdout NSP for Assault, Skiing, and Montezuma, while PoE-World remains stronger on Alien random replay.

\begin{table}[!t]
  \centering
  \scriptsize
  \setlength{\tabcolsep}{2.5pt}
  \renewcommand{\arraystretch}{1.02}
  \begin{tabular}{@{}lcccccc@{}}
    \toprule
    Game
      & \multicolumn{3}{c}{Chosen-action NSP}
      & \multicolumn{3}{c}{Branch NSP} \\
    \cmidrule(lr){2-4}\cmidrule(lr){5-7}
     & \shortstack{LLM-as\\WM}
     & \shortstack{PoE-\\World}
     & \shortstack{\textbf{Mind-Studio}\\\textbf{(ours)}}
     & \shortstack{LLM-as\\WM}
     & \shortstack{PoE-\\World}
     & \shortstack{\textbf{Mind-Studio}\\\textbf{(ours)}} \\
    \midrule
    Alien     & \textbf{0.67} & 0.65 & \textbf{0.67} & 0.22 & \textbf{0.35} & 0.22 \\
    Assault   & 0.00 & 1.00 & \textbf{1.00} & 0.00 & 0.33 & \textbf{1.00} \\
    Skiing    & \textbf{1.00} & 0.97 & \textbf{1.00} & 0.51 & 0.80 & \textbf{0.83} \\
    Montezuma & 0.42 & 0.42 & \textbf{0.51} & \textbf{0.27} & 0.12 & 0.26 \\
    \bottomrule
  \end{tabular}
  \caption{\textbf{Offline NSP on replay states (Qwen3-Coder-480B).} Train-session states; no planner in the loop. Per-game lookahead horizon $K$ in Table~\ref{tab:app_k}.}
  \label{tab:offline_replay_nsp}
\end{table}

\begin{table}[!t]
  \centering
  \scriptsize
  \setlength{\tabcolsep}{2.5pt}
  \renewcommand{\arraystretch}{1.02}
  \begin{tabular}{@{}lcccccc@{}}
    \toprule
    Game
      & \multicolumn{3}{c}{Chosen-action NSP}
      & \multicolumn{3}{c}{Branch NSP} \\
    \cmidrule(lr){2-4}\cmidrule(lr){5-7}
     & \shortstack{LLM-as\\WM}
     & \shortstack{PoE-\\World}
     & \shortstack{\textbf{Mind-Studio}\\\textbf{(ours)}}
     & \shortstack{LLM-as\\WM}
     & \shortstack{PoE-\\World}
     & \shortstack{\textbf{Mind-Studio}\\\textbf{(ours)}} \\
    \midrule
    Alien     & 0.16 & \textbf{0.23} & 0.16 & 0.15 & \textbf{0.24} & 0.15 \\
    Assault   & 0.00 & 0.33 & \textbf{0.98} & 0.00 & 0.33 & \textbf{0.98} \\
    Skiing    & 0.31 & 0.66 & \textbf{0.69} & 0.37 & 0.67 & \textbf{0.70} \\
    Montezuma & 0.12 & 0.05 & \textbf{0.16} & 0.10 & 0.03 & \textbf{0.12} \\
    \bottomrule
  \end{tabular}
  \caption{\textbf{Offline NSP on random states (Qwen3-Coder-480B).} Test-session states; no planner in the loop. Per-game lookahead horizon $K$ in Table~\ref{tab:app_k}.}
  \label{tab:offline_random_nsp}
\end{table}

\paragraph{Online fidelity.}
Inside LLM-as-Planner, each planning round snapshots the live ALE state, rolls out every candidate action for $K$ frames under both the WM and Real ALE ($K{=}16$ on Skiing, $K{=}8$ elsewhere), and counts an exact hit when the final \texttt{player\_xy} matches pixel-wise. Chosen-action NSP tests only the committed action; branch NSP pools all candidate actions. Thus, the NSP columns in Table~\ref{tab:main_results} measure fidelity on the states actually visited by the planner.

\paragraph{LLM-as-WM failure mode.}
LLM-as-WM must infer $K$-frame geometry from text rather than execute transition code, which explains its near-zero exact NSP on movement-heavy cases in Table~\ref{tab:main_results}. 

\subsection{Planning Results}
\label{sec:planning_results}
Table~\ref{tab:main_results} is the planning results grid using the world model predictions. Each row fixes the planner and changes only the lookahead source; \textsc{Real-ALE} is an oracle reference, not a learned WM. The Qwen3-Coder-480B row gives the cleanest source comparison: Mind-Studio matches the best learned task outcome on 4 games, improves the chosen-action NSP and branch action NSP on most cases.

The non-Qwen rows test whether the same previews remain useful across planner backbones. Assault is the most stable fidelity case: Mind-Studio has the strongest chosen-action and branch NSP under all five planners, and obtains the best or tied-best learned task outcome under $3/5$ planners. Alien and Montezuma vary more by planner, and Skiing is a weak result: only the Qwen3-Coder-480B planner obtains nonzero task progress, while other planners match the zero-score baselines and branch NSP often ties or only slightly exceeds PoE-World (Table~\ref{tab:cross_planner_fidelity}). On Montezuma, the Qwen3-Coder-480B row reaches $5/8$ verified subgoals under Mind-Studio, PoE-World, and LLM-as-WM alike (Table~\ref{tab:main_results}), indicating that lookahead fidelity rather than subgoal count separates the sources at this horizon.

\subsection{Per-Action Fidelity}
\label{sec:action_fidelity}

Table~\ref{tab:action_precision} breaks pooled branch NSP into per-action rates, exposing which actions drive the aggregate differences.

\begin{table}[!t]
  \centering
  \scriptsize
  \setlength{\tabcolsep}{1.5pt}
  \renewcommand{\arraystretch}{0.95}
  \begin{tabular}{@{}llccc@{}}
    \toprule
    \textbf{Game} & \textbf{Action}
      & \shortstack{LLM-as\\WM}
      & \shortstack{PoE-\\World}
      & \shortstack{\textbf{Mind-Studio}\\\textbf{(ours)}} \\
    \midrule
    \multirow{3}{*}{\shortstack{Skiing\\($K{=}16$)}}
      & \textsc{Noop}  & 0.15 & 0.17 & \textbf{0.18} \\
      & \textsc{Left}  & 0.23 & \textbf{0.25} & 0.09 \\
      & \textsc{Right} & 0.00 & 0.08 & \textbf{0.18} \\
    \midrule
    \multirow{3}{*}{\shortstack{Assault\\($K{=}8$)}}
      & \textsc{Up}    & 0.00 & \textbf{1.00} & \textbf{1.00} \\
      & \textsc{Left}  & 0.00 & 0.00 & \textbf{1.00} \\
      & \textsc{Right} & 0.00 & 0.32 & \textbf{0.61} \\
    \midrule
    \multirow{8}{*}{\shortstack{Alien\\($K{=}8$)}}
      & \textsc{Noop}  & 0.84 & \textbf{1.00} & 0.65 \\
      & \textsc{Fire}  & 0.84 & \textbf{1.00} & 0.65 \\
      & \textsc{Up}    & 0.00 & \textbf{1.00} & 0.65 \\
      & \textsc{Down}  & 0.00 & 0.00 & \textbf{0.25} \\
      & \textsc{Left}  & 0.00 & 0.17 & \textbf{0.55} \\
      & \textsc{Right} & 0.00 & \textbf{0.17} & 0.15 \\
      & \textsc{LFire} & 0.42 & 0.17 & \textbf{0.55} \\
      & \textsc{RFire} & 0.05 & \textbf{0.17} & 0.15 \\
    \midrule
    \multirow{6}{*}{\shortstack{Montezuma\\($K{=}8$)}}
      & \textsc{Noop}  & \textbf{0.99} & 0.50 & \textbf{0.99} \\
      & \textsc{Fire}  & \textbf{0.99} & 0.00 & 0.00 \\
      & \textsc{Up}    & 0.01 & \textbf{0.49} & 0.01 \\
      & \textsc{Down}  & \textbf{0.48} & 0.00 & 0.00 \\
      & \textsc{Left}  & 0.00 & 0.00 & \textbf{0.48} \\
      & \textsc{Right} & 0.00 & 0.01 & \textbf{0.49} \\
    \bottomrule
  \end{tabular}
  \caption{\textbf{Per-action branch NSP (Qwen3-Coder-480B).} Bold = best among learned sources. Alien \textsc{noop}/\textsc{fire} are diagnostic and excluded from pooled branch NSP.}
  \label{tab:action_precision}
\end{table}

On Montezuma, LLM-as-WM's branch NSP is inflated by stationary actions (\textsc{noop}/\textsc{fire} near $1.00$ and \textsc{down} at $0.48$) while remaining at $0.00$ on \textsc{left}/\textsc{right}; Mind-Studio has the reverse profile, with stronger lateral movement but weaker stationary-action fidelity.

\paragraph{NSP--task-success relationship.}
Across $56$ learned-source pairs, higher chosen-action NSP ties or improves task outcome in $44$ cases (Sec.~\ref{sec:backbone}), confirming NSP as a useful proxy. The gap between NSP and SR stems from two sources. First, the synthesized WM is \emph{deterministic}: it cannot model stochastic respawns or random enemy motion, so an action that looks safe in preview may still fail in the real environment. Second, a one-frame positional error compounds over $K$ frames, producing large endpoint discrepancies even when the first-step prediction is nearly correct. Alien egg counts and Skiing's momentum-overshoot are the clearest examples; see Sec.~\ref{sec:backbone} for detail.

\subsection{Ablation: synthesis inputs}
\label{sec:ablation}

We ablate synthesis inputs on Montezuma, the game with the most complex rule set. All synthesizers use a 128K context window and are scored by one-step holdout NSP on \texttt{selected.py} over the random-policy session (988 transitions), using the same exact-pixel \texttt{player\_xy} precision metric as planning-time NSP. This offline metric is easier than $K$-step planning NSP, but it isolates whether the synthesized transition program generalizes beyond the replay trace used for prompting.

In the reported table, Default uses 279 entropy-selected compact rows, while $-$compression uses 17 consecutive full $(s_t,a_t,s_{t+1})$ prefixes---about $16.4\times$ ($\approx$1540\%) more interaction examples within the same context budget. Removing compression sharply lowers exact holdout NSP for every synthesizer. The $-$setting ablation also hurts the Qwen models, while DeepSeek is less sensitive. Replacing replay traces with random-policy data helps Qwen3-Coder-480B under exact match ($0.52$ vs.\ $0.46$) but not DeepSeek ($0.12$ vs.\ $0.31$).

\begin{table}[!t]
  \centering
  \scriptsize
  \setlength{\tabcolsep}{2.5pt}
  \renewcommand{\arraystretch}{1.05}
  \begin{tabular}{@{}lcccc@{}}
    \toprule
    \multirow{2}{*}{Synthesizer}
      & \multicolumn{3}{c}{Ablated input}
      & \multirow{2}{*}{\shortstack{Default\\$\widetilde{\mathcal{T}}$}} \\
    \cmidrule(lr){2-4}
      & \shortstack{$-$data\\random}
      & \shortstack{$-$comp.\\full $(s,a,s')$}
      & \shortstack{$-$setting\\no skill}
      & \\
    \midrule
    Qwen3-Coder-480B
      & \textbf{0.521} & 0.077 & 0.244 & 0.462 \\
    Qwen3-Max-Thinking
      & 0.219 & 0.187 & 0.137 & \textbf{0.352} \\
    DeepSeek-V4-Flash
      & 0.119 & 0.035 & 0.298 & \textbf{0.313} \\
    \bottomrule
  \end{tabular}
  \caption{\textbf{Synthesis-input ablation} on Montezuma held-out one-step NSP (exact-pixel \texttt{player\_xy} precision).}
  \label{tab:step2_ablation}
\end{table}

\subsection{Backbone Stability and Task Alignment}
\label{sec:backbone}
Table~\ref{tab:main_results} sweeps five planner LLMs while holding each selected Mind-Studio program fixed, and Table~\ref{tab:cross_planner_fidelity} averages branch fidelity across those planners. Higher fidelity usually helps: across $56$ learned-source pairs with different chosen-action NSP (three learned sources compared within each fixed game--planner row, omitting NSP ties), the higher-NSP source ties or improves task outcome in $44$ cases and strictly improves it in $16$; branch NSP gives a similar pattern ($43$ ties-or-better, $16$ strict wins). Still, the exceptions matter. Alien shows that aggregate NSP can reward easy \textsc{noop}/\textsc{fire} predictions while missing lateral movement, so task outcome and per-action coverage must be read together.
The sweep reinforces the game-level pattern. Assault is robust in fidelity, with Mind-Studio much higher than PoE-World in mean branch NSP ($0.91$ vs.\ $0.36$) and best or tied-best learned task outcome under $3/5$ planners. Skiing remains weak despite a small fidelity edge ($0.27$ vs.\ $0.24$): offline NSP on Skiing is strong, but holding \textsc{left} for $K{=}16$ frames causes over-steering that reduces downward speed to near zero; the $K$-hold protocol thus cannot capture the cumulative directional cost of sustained holds, making Skiing hard for any bounded-lookahead planner. Montezuma and Alien depend more strongly on the planner backbone.

\begin{table}[t]
  \centering
  \small
  \setlength{\tabcolsep}{5pt}
  \resizebox{\linewidth}{!}{%
  \begin{tabular}{@{}lccc@{}}
    \toprule
    Game & \textbf{Mind-Studio} & PoE-World & LLM-as-WM \\
    \midrule
    Skiing ($K{=}16$)              & \textbf{0.266} & 0.239 & 0.046 \\
    Assault ($K{=}8$)              & \textbf{0.912} & 0.361 & 0.069 \\
    Alien ($K{=}8$)\textsuperscript{$\ddagger$}  & \textbf{0.338} & 0.284          & 0.024 \\
    Montezuma's Revenge ($K{=}8$)  & 0.186          & 0.093          & \textbf{0.215} \\
    \bottomrule
  \end{tabular}
  }
  \caption{\textbf{Cross-planner branch fidelity.} Each cell is the mean exact-pixel branch NSP across the five planner LLMs in Table~\ref{tab:main_results}. \textsuperscript{$\ddagger$}Alien omits \textsc{noop}/\textsc{fire}.}
  \label{tab:cross_planner_fidelity}
\end{table}

\section{Conclusion}

Mind-Studio turns an agent's internal ``image of the world'' into a runnable program. The same program serves both as a per-action next-state predictor and as the lookahead source for an LLM planner with subgoal rollback. This composition mirrors how a player approaches an unfamiliar environment: infer rules, simulate candidate actions, commit, and back up after failure. On Montezuma's Revenge and three respawn-driven OC-Atari games, the synthesized WM improves per-action accuracy over an ensemble baseline and lets LLM-as-Planner reach more subgoals than BFS within an order-of-magnitude smaller wall-time budget, with planner-backbone sweeps exposing where task outcomes remain sensitive to planner choice.

More broadly, Mind-Studio demonstrates that a single, unrefined LLM pass over entropy-selected symbolic traces is sufficient to produce an executable world model that meaningfully supports planning in partially observable settings. The game skill file, which can be seeded from pixel-level scene extraction alone, bridges the gap between raw visual observations and the structured dynamics knowledge that an LLM synthesizer needs. We believe this design---skill-file-guided single-pass synthesis evaluated by $K$-step lookahead fidelity---is a reusable blueprint for building lightweight, interpretable world models in any domain where object-level observations and discrete actions are available.

\section*{Limitations}

Mind-Studio's remaining errors are concentrated in geometry-sensitive contexts (compound rope/skull interactions, ladder--platform overlap), pointing to a clear improvement path: targeted trace collection around rare interaction types together with richer transition templates for residual stochastic cases. Task success also depends on exploration quality: Alien egg counts still vary across planner backbones despite strong per-action NSP, indicating that route-discovery prompts are complementary to better world models.

LLM-as-WM's near-zero movement NSP suggests an alternative direction: extracting explicit transition rules---conditions, priority orderings, context flags---and injecting them as structured context into the LLM, potentially matching the hidden-rule coverage of executable code without a compiled program. Our experiments are also scoped to Atari 2600 with $\le 20$ object types and an 8-action space; extending to richer vocabularies or continuous control is a natural next step.

\section*{Ethical Considerations}

This work uses publicly available Atari 2600 ROMs and the open-source OC-Atari extractor; no human-subject data or sensitive content is involved. The synthesized world models are restricted to game dynamics and have no foreseeable misuse profile beyond the underlying LLM and ALE software. Released artifacts (programs, prompts, and logs) are intended for academic reproducibility. Applications beyond games, especially in higher-stakes domains such as robotics, healthcare, or finance, should pair executable model synthesis with domain-specific validation and safety checks.

\section*{Acknowledgments and AI Assistance}

In accordance with ARR/ACL disclosure guidance, we used AI tools for limited
editorial support, specifically grammar polishing and syntax-error correction in
manuscript text and code snippets. All research design, experiments, numerical
results, interpretations, and final wording decisions were made and verified by
the authors.

\bibliography{anthology}
\begin{appendix}

\section{Subgoal Curriculum and Predicate Specification}
\label{sec:app_subgoal}

We factorize Montezuma's Revenge into $K{=}8$ ordered milestones. The agent must complete them in sequence, and no credit is given for out-of-order completions.

\begin{enumerate}
  \item \textbf{Reach ladder foot.} Agent's bottom edge within 8\,px of the base tile of the first ladder.
  \item \textbf{Climb ladder.} Agent's $y$-coordinate passes the midpoint of the ladder segment.
  \item \textbf{Cross rope.} Agent collides with the rope sprite and translates horizontally by $\ge 16$\,px while the rope-contact flag is set.
  \item \textbf{Descend to key room.} Agent's bounding box enters the key-room region (fixed pixel rectangle, empirically determined).
  \item \textbf{Obtain key.} The key-object sprite disappears from the scene, and agent inventory count increments.
  \item \textbf{Return to door.} Agent's bounding box overlaps the door sprite in the key room.
  \item \textbf{Unlock door.} Inventory count decrements; the door-open animation flag becomes active.
  \item \textbf{Reach room exit.} Agent exits the current screen region through the designated portal.
\end{enumerate}

Each predicate $\Phi_k$ is a Python function that reads axis-aligned bounding boxes directly from the \texttt{OC-Atari} object extractor~\cite{ocatari}. All spatial thresholds are fixed at evaluation time and are identical across world-model conditions. The environment maintains a monotonic counter $m$, which increments from $m$ to $m{+}1$ only when $\Phi_{m+1}$ fires on the \emph{true} game state returned by ALE, never on a WM prediction.

\begin{algorithm}[ht]
\small
\caption{LLM-as-Planner with subgoal rollback}
\label{alg:llm-planner}
\begin{algorithmic}[1]
\Require WM $M^\star$, LLM policy $\pi_{\text{LLM}}$, predicates $\{\Phi_j\}_{j=0}^{J}$, env.\ $\mathcal{E}$, budget $T_{\max}$, stuck cap $N_{\text{stuck}}$
\State $m\!\gets\!0$; $s_0\!\gets\!\text{reset}(\mathcal{E})$; $s^{\dagger}_{0}\!\gets\!s_0$; $\text{stuck}\!\gets\!0$
\For{$t=0,\dots,T_{\max}-1$}
  \For{$a\in\mathcal{A}$} \Comment{$K$-frame lookahead via $M^\star$}
    \State $\hat{s}_{t,a}\!\gets\!(M^\star)^K(s_t,a)$
  \EndFor
  \State $B_t\!\gets\!\{\hat{s}_{t,a}\}_{a\in\mathcal{A}}$
  \State $a_t\!\sim\!\pi_{\text{LLM}}(\cdot\mid s_t,B_t,m)$
  \State $s_{t+1}\!\gets\!\mathcal{E}.\text{step}(a_t)$
  \If{$\Phi_{m+1}(s_{t+1})=1$} \Comment{subgoal verified}
    \State $m\!\gets\!m+1$; $s^{\dagger}_{m}\!\gets\!s_{t+1}$; $\text{stuck}\!\gets\!0$
  \ElsIf{$\text{Reset}(s_{t+1})$ \textbf{or} $\text{stuck}\!\ge\!N_{\text{stuck}}$}
    \State $s_{t+1}\!\gets\!\mathcal{E}.\text{restore}(s^{\dagger}_{m})$; $\text{stuck}\!\gets\!0$ \Comment{rollback}
  \Else
    \State $\text{stuck}\!\gets\!\text{stuck}+1$
  \EndIf
  \If{$m=J$} \textbf{break} \EndIf
\EndFor
\State \Return verified prefix $m$, trajectory
\end{algorithmic}
\end{algorithm}

\section{Implementation Details}
\label{sec:app_impl}

\subsection{Models and decoding}
All LLMs are accessed through their commercial inference APIs with decoding temperature $0$ for code synthesis, repair, and planning. The exact identifiers used in the reported runs are:
\begin{itemize}\itemsep0pt
  \item \textsc{Qwen3-Coder-480B-A35B-Instruct-Turbo} -- WM synthesizer (default) and main planner;
  \item \textsc{Qwen3-Max-Thinking} -- WM synthesizer ablation;
  \item \textsc{DeepSeek-V4-Pro} (synth), \textsc{DeepSeek-V4-Flash} (planner);
  \item \textsc{Gemini-2.5-Flash} -- WM synthesizer ablation and planner stability sweep;
  \item \textsc{Claude-4-Sonnet} -- WM synthesizer ablation and planner stability sweep;
  \item \textsc{Llama-4-Scout-17B-16E} -- WM synthesizer ablation and planner stability sweep.
\end{itemize}

\subsection{Synthesis pipeline}
The synthesizer receives (i)~the entropy-selected subset $\widetilde{\mathcal{T}}$ (see Sec.~\ref{sec:entropy_selection} for the score and Sec.~\ref{sec:wm_synthesis} for the assembly rule), (ii)~a static-map / rule list induced from $\mathcal{D}$ for the \emph{all} branch, and (iii)~a physics-summary block. The token budget per synthesis call is $32{,}768$ input / $8{,}192$ output. The refinement loop runs for at most $3$ iterations. Each iteration replays $\mathcal{D}_{\text{val}}$, returns up to $20$ mismatch rows, and asks the synthesizer for a local patch.

\subsection{Entropy-score hyperparameters}
For Eq.~\eqref{eq:entropy-score}, ordinary movement uses
$\omega_{\text{move}}=1$, $\alpha=0.3$, and $\beta=0.5$, so a position update
contributes $1+0.3|dx|+0.5|dy|$. Appearance and disappearance changes in
$\omega_{\text{change}}(o)$ contribute $10$. The relation-change and explicit-event
weights are $\omega_{\text{rel}}=4$ and $\omega_{\text{evt}}=10$, respectively.
These constants are fixed across all games; only optional coverage constraints
for rare observed interactions are game-specific.

\subsection{Planning harness}
Algorithm~\ref{alg:llm-planner} is run with $T_{\max}=300$ steps on Montezuma's Revenge and Alien, $T_{\max}=200$ on Skiing, and $T_{\max}=100$ on Assault. The stuck cap is $N_{\text{stuck}}=12$. The commit-chain length $K$ is game-specific (Table~\ref{tab:app_k}), matching the OC-Atari frameskip used to record the training traces. Each LLM call returns a single action or a length-$K$ chunk; chunked outputs are truncated if a checkpoint commit fires mid-chunk. All methods use the same evaluation configuration within each game and planner condition.

\begin{table}[h]
  \centering
  \small
  \setlength{\tabcolsep}{6pt}
  \begin{tabular}{@{}lc@{}}
    \toprule
    Game & Lookahead horizon $K$ \\
    \midrule
    Alien & 8 \\
    Assault & 8 \\
    Skiing & 16 \\
    Montezuma's Revenge & 8 \\
    \bottomrule
  \end{tabular}
  \caption{Per-game lookahead horizon used in offline and online NSP evaluation.}
  \label{tab:app_k}
\end{table}

\subsection{Computational resources}
WM synthesis is API-only. Each game's full pipeline (entropy selection, synthesis, patching, optional refinement, and single-step evaluation) consumes $\le 200$ LLM calls and finishes in under $20$ minutes wall clock per game on a workstation that drives only the API and the OC-Atari sandbox. Planning runs are CPU-only (ALE + OC-Atari + LLM API); a full $300$-round Montezuma planning episode under Mind-Studio finishes in $\sim 7\,\text{s}$ on a single core. No GPU is required for any reported number.

\subsection{Reproducibility}
We release: (i)~the synthesized WM programs ($M^{\star}$ per game) used in all reported tables; (ii)~the prompt templates for synthesis, refinement, and planning; (iii)~per-game subgoal predicates ($\Phi_k$, Appendix~\ref{sec:app_subgoal}); (iv)~the raw planning logs behind every cell in Tables~\ref{tab:main_results} and~\ref{tab:cross_planner_fidelity}; and (v)~the entropy-selector code (Eq.~\eqref{eq:entropy-score}) and WM evaluator. All runs use the OC-Atari extractor at its default configuration~\cite{ocatari} and the public Atari 2600 ROMs.

\section{Game setting skill-file excerpt}
\label{sec:app_game_setting_excerpt}

Table~\ref{tab:step2_ablation} uses a game setting / skill block during WM synthesis in the Default condition. The following is an excerpt from
\texttt{games/montezumarevenge/ prompt\_settings\_en.txt}, showing the concrete content that is removed in the $-$setting ablation.

{\footnotesize
\begin{verbatim}
[game_setting]
Use up/down/left/right and FIRE-like actions
to move and jump; on ropes/ladders you can
climb up/down; touching a skull kills and
restarts; the goal is to get the key.

[patch_actions]
"NOOP", "LEFT", "RIGHT", "UP", "DOWN",
"FIRE", "LEFTFIRE", "RIGHTFIRE"
\end{verbatim}
}

\end{appendix}

\end{document}